# Path Planning and Navigation Inside Off-World Lava Tubes and Caves

Himangshu Kalita
Aerospace and Mechanical
Engineering Department
University of Arizona
Tucson, AZ 85721
hkalita@email.arizona.edu

Steven Morad
Aerospace and Mechanical
Engineering Department
University of Arizona
Tucson, AZ 85721
smorad@email.arizona.edu

Aaditya Ravindran
School of Electrical, Computer and
Energy Engineering
Arizona State University
Tempe, AZ 85287
aravin11@asu.edu

Jekan Thangavelautham
Aerospace and Mechanical
Engineering Department
University of Arizona
Tucson, AZ 85721
jekan@email.arizona.edu

*Abstract*—Detailed surface images of the Moon and Mars reveal hundreds of cave-like openings. These cave-like openings are theorized to be remnants of lava-tubes and their interior maybe in pristine conditions. These locations may have well preserved geological records of the Moon and Mars, including evidence of past water flow and habitability. Exploration of these caves using wheeled rovers remains a daunting challenge. These caves are likely to have entrances with caved-in ceilings much like the lava-tubes of Arizona and New Mexico. Thus, the entrances are nearly impossible to traverse even for experienced human hikers. Our approach is to utilize the SphereX robot, a 3 kg, 30 cm diameter robot with computer hardware and sensors of a smartphone attached to rocket thrusters. Each SphereX robot can hop, roll or fly short distances in low gravity, airless or low-pressure environments. Several SphereX robots maybe deployed to minimize single-point failure and exploit cooperative behaviors to traverse the cave. There are some important challenges for navigation and path planning in these cave environments. Localization systems such as GPS are not available nor are they easy to install due to the signal blockage from the rocks. These caves are too dark and too large for conventional sensor such as cameras and miniature laser sensors to perform detailed mapping and navigation. In this paper, we identify new techniques to map these caves by performing localized, cooperative mapping and navigation. In our approach, a team of SphereX robots much like a team of cave explorer will adopt specialized roles to perform navigation. For a minimal science mission, these robots need to obtain camera images and basic maps of the cave interior to be transmitted back to a lander or rover situated outside the cave. The teams of SphereX robots form a bucket brigade and partition the currently accessible volume of the cave. Then the teams of robots attempt to expand their reach deeper into the cave and sense their progress. Imaging the cave interior is expensive and require use of high-power strobe lights. The images would be compiled into a 3D point cloud and meshed by the lander or transmitted to ground. Using this conservative approach, we ensure the robots are always within communication reach of a lander/rover outside the cave. Once large segments of the cave are mapped, the rovers may lay down a network of mirrors to beam sunlight and laser light from a base station at the cave entrance to the far reaches of the cave. These mirrors also help the robots identify a pathway back to the cave entrance. Efforts are underway to perform field experiments to validate the feasibility our proposed approach to cave exploration.

## I. Introduction

The latest orbital images of the Moon and Mars taken by the Lunar Reconnaissance Orbiter (LRO) and Mars Reconnaissance Orbiter (MRO) respectively reveal hundreds of cave openings known as pits (Fig. 1) [37]. These pits serve as natural shelters from micro-meteoroids, cosmic radiation, and surface temperature extremes. Some have proposed these pits could be ideal for setting up a subsurface human base.

Mobile ground robots have become integral for surface exploration of the Moon, Mars and other planetary bodies. These rovers have proven their merit, but they are large, in the order of several hundred kilograms and house state-of-the-art science laboratories. Exploring these pits using wheeled ground robots will prove to be a daunting challenge. Many pits from LRO images show evidence of collapsed entrances due to past geologic activity. On Earth, many lava-tubes have collapsed entrances that are extremely rugged and are only accessible by some of the fittest hikers. In addition, conventional methods of path-planning and navigation used by planetary rovers are not applicable in these pits as they are sheltered by thick rock, which block sunlight and prevent radio communication with the outside world.

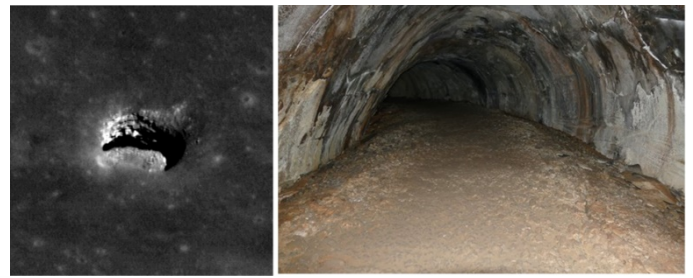

Fig. 1. Hundreds of pits such as Mare Ingenii (left) have been found on the Moon by the Lunar Reconnaisance Orbiter (LRO). These pits are thought to be lava tubes simillar to Lava River Cave in Flagstaff, Arizona (right).

Exploring these pits requires a whole new mobility platform and specialized methods for path planning and navigation. In this paper, we present a spherical robot platform called SphereX with mass of 3 kg and diameter of 30 cm and that can fly, hop and roll. As we show through simulations, a team of SphereX robots can cooperatively map, path plan and navigate through a lava-tube. SphereX's flight capabilities are intended to match that of a terrestrial quad-copter drone but do so in low-gravity, air-less or low-pressure environments. SphereX will contain electronics and sensors equivalent to current smartphones attached to rocket thrusters [2],[3],[4]. Each robot also contains an array of guidance, navigation and control sensors and volume for a 1 kg science payload. Flying, especially hovering consumes significant fuel and hence will be used sparingly.

We have developed alternative mobility solutions for SphereX that improves on fuel use and range. Ballistic hops overcome obstacles that maybe many times larger than the robot, enabling short flights, while also providing range. To explore a cave or lava tube requires a team of SphereX robots that work cooperatively to map, navigate and communicate the data back to a base station located outside the lava-tube. Added to the challenge, the robots also need to perform localization in the dark. A lack of line-of-sight communication and interference from the thick rock requires the robots operate in bucket brigades, as they relay messages from the base station to individual robots. Using this approach, we propose the SphereX robots install small mirrors along the length of the lava tube much like a "trail of breadcrumbs." This trail of mirrors would simplify localization and enable the robots to easily travel to the lava-tube entrance or to the tube end. The mirrors would be used to beam sunlight along the network of mirrors, providing lighting [36]. An additional laser beam would be reflected using these mirrors to beam power [34] and provide for a fixed communication line [35].

Our approach to exploration shows that with a team of robots, it may be possible to scale-up to cover large areas in short duration [1]. Multiple robots operating as a team offer significant benefits over a single large rover, as they are not prone to single-point failure, enable distributed command and control and enable exploration in parallel. In addition, these relatively low-cost robots can complement large, high-value rovers and landers, helping to explore inaccessible, high-risk, high-reward sites, without risking the overall mission. In the following sections, we present background and related work, followed by a system overview of the SphereX robot, presentation on our multirobot path-planning and navigation algorithm followed by discussions, conclusions and future-work.

## II. RELATED WORK AND MOTIVATION

Small spherical robots have been widely proposed in the past. Their spherical shape enables them to roll on loose, even terrain. Examples include spherical robots developed at Univ. of Sherbrooke [5], Kickbot [6] developed at MIT, Cyclops [7] at Carnegie Mellon University and inflatable ball robots developed at North Carolina State University [8] and University of Toronto [9]. Typically, these spherical robots use a pair of direct drive motors in a holonomic configuration. Others such as the Cyclops and the inflatables pivot a heavy mass, thus moving center of gravity that results in rolling. Other mobility techniques including use of spinning flywheels attached to a two-link manipulator on the Gyrover [10] or 3-axis reaction wheels to spin and summersault as with the Hedgehog developed by Stanford and NASA JPL [11]. Hedgehog's use of reaction wheels enables it to overcome rugged terrain by simply creeping over the obstacle no matter how steep or uneven. However, it's unclear if a gyro-based system can overcome both steep and large obstacles. In reality, even a gyro-based system is bound to slip on steep surfaces, but under low gravity environments such as asteroids, they may be able to reach meters in height.

An alternative to rolling and creeping is hopping. A typical approach to hopping is to use a hopping spring mechanism to overcome large obstacles [12]. One is the Micro-hopper for Mars exploration developed by the Canadian Space Agency [13]. The Micro-hopper has a regular tetrahedron geometry that enables it to land in any orientation at the end of a jump. The hopping mechanism is based on a novel cylindrical scissor mechanism enabled by a Shape Memory Alloy (SMA) actuator. However, the design allows only one jump per day on Mars. Another technique for hopping developed by Plante and Dubowsky at MIT utilize Polymer Actuator Membranes (PAM) to load a spring. The system is only 18 grams and can enable hopping of Microbots with a mass of 100 g up to 1 m [14],[15]. Microbots are cm-scale spherical robots equipped with power and communication systems, a mobility system that enables it to hop, roll and bounce and an array of miniaturized sensors such as imagers, spectrometers, and chemical analysis sensors developed at MIT. They are intended to explore caves, lava-tubes, canyons and cliffs. Ideally, many hundreds of these robots would be deployed enabling large-scale in-situ exploration.

Many algorithms have also been developed to explore unknown new environments. One of the most popular one is the Sensor-based Random Tree (SRT) navigation tree [16]. The algorithm is based on incremental construction of a tree-type data structure through the random generation of robot configurations within a local security (LSR) area detected by the robot sensors where the robot can move without risk of collision with an obstacle. Another algorithm based on the SRT algorithm for a multi robot case is the Sensor-based Random Graph (SRG) algorithm [17]. In this algorithm, the tree structure is transformed into an exploration graph when a safe route to travel between two nodes is found. The random exploration graph (REG) is another SRT based algorithm, in which the tree structure becomes an exploration graph when a safe route is found to travel between two nodes without hierarchical relationship, which is determined based on the intersection of the free frontiers belonging to these nodes [18]. Also, for path planning and exploration, the environment can also be modelled in different ways. One way is to model as a geometric structure with polygonal obstacles where it assumes that the robot knows everything within line-of-sight visibility [19]. Another way is to model it as a grid in which some cells are open, others are blocked, and some are unknown or more complicated cell states [20]. One more way is to model the environment as a graph, nodes corresponding to locations, and edges corresponding to passages between the locations [21]. However, most of the algorithms are confined to wheeled robots and doesn't address their approach for multiple hopping robots in an unknown environment.

SphereX is the direct descendant of the Microbot platform. SphereX has the same goals as the Microbots, but with the goal of launching fewer robots, that are better equipped with science-grade instruments. Moreover, the path planning approach addressed in this paper is motivated by multiple hopping robots navigating a maze [38], where one robot hops at a time within a local area of safety, collects more information and plans for their next hop. Our past work has shown has shown the feasibility of multiple small robots working together as a network [25],[26],[27],[28].

## III. SYSTEM OVERVIEW

In this section, we present the SphereX spherical robot that is capable of hopping, flying and rolling through caves, lava-tubes and skylights. Fig. 2 shows the internal and external views of each SphereX robot. The lower half of the sphere contains the power and propulsion system, with storage tanks for fuel and oxidizer connected to the main thruster. It also contains a 3-axis reaction wheel system for maintaining roll, pitch and yaw angles and angular velocities along *x*, *y* and *z* axes. The propulsion unit provides thrust along +z axis and the reaction wheel system control the attitude and angular velocity of the robot that enables it to perform ballistic hop. Next is the Lithium Thionyl Chloride batteries with specific energy of 500 Wh/kg arranged in a circle. An alternative to batteries are PEM fuel cells. PEM fuel cells are especially compelling as techniques have been developed to achieve high specific energy using solid-state fuel storage systems that promise 2,000 Wh/kg [22] [23]. However, PEM fuel cells require development for a field system in contrast to lithium thionyl chloride that has already been demonstrated on deep space.

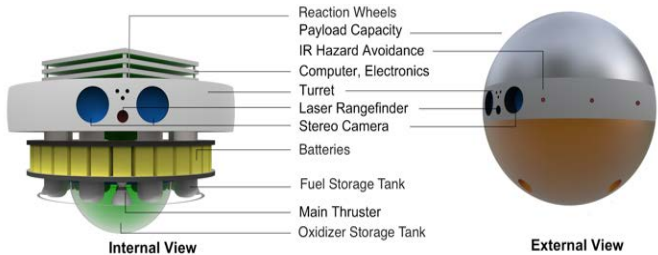

Fig. 2. Internal and External view of the SphereX robot.

For sensing, planning and control, a pair of stereo cameras and a laser range finder is mounted to each robot and they roll on a turret. This enables the robot to take panoramic pictures and scan the environment without having to move using the propulsion system. Above the turret are two computer boards, IMU and IO-expansion boards, in addition to a power board. The volume above the electronics is reserved for payload of up to 1 kg. Apart from the proposed propulsion subsystem, all the other hardware components can be readily assembled using Commercial off-the-self (COTS) components.

## IV. BALLISTIC HOPPING

### A. Simplified model and calculation of initial velocity

Neglecting the effect of irregular gravity field on a target body, a single hopping motion can be simplified as a parabolic motion. The robot needs to hop from rest position $r_{t0}$ with velocity $v_{t0}$ and impact position $r_{tf}$ with velocity $v_{tf}$. The corresponding initial velocity $v_{t0}$ needed to move the robot from its initial position to its final position can be computed based on its parabolic motion. Considering $d = d_x \hat{\imath} + d_y \hat{\jmath}$ as the vector connecting the initial point and the final point, $g$ as the acceleration due to gravity vector and $\tau$ as the transfer time, the components of the initial velocity can be computed as $v_z = g\tau/2$, $v_x = d_x/\tau$ and $v_y = d_y/\tau$. Hence $v_{t0}$ can be expressed as

$$v_{t0} = v_x\hat{\imath} + v_y\hat{\jmath} + v_z\hat{k} \qquad (1)$$

So, for pinpoint landing the robot must apply an initial impulsive thrust to achieve initial delta *v* of $\Delta v_1 = |v_{t0}|$ and for soft landing it has to apply another impulsive thrust to achieve a final delta v of $\Delta v_2 = |-v_{tf}|$.

### B. Trajectory optimization

For pinpoint soft landing two impulses are applied for each hopping trajectory. Considering a single hopping movement, the optimization objective is to minimize the fuel consumption and the optimal index can be expressed by

$$J = \int_0^\tau \|T\| \, dt \qquad (2)$$

For 1 kg of propellant with a specific impulse (Isp) equal to 350 s, the distance travelled per hop has a huge impact on the number of hops possible and the total distance covered. Fig. 3 shows the variation of number of hops and total distance covered with varying single hopping distance with optimized fuel consumption on the surface of Moon and Mars. It can be seen that on the surface of the Moon, with a 1 m hopping distance, the robot can perform 546 hops and hence can travel 546 m. However, if the hopping distance is increased to 100 m, it can perform 55 hops only but the total hopping distance increases to 5500 m. In the case of Mars, with 1 m hopping distance, it can cover a distance of 359 m and with 100 m hopping distance, the total hopping distance increases to 4,100 m.

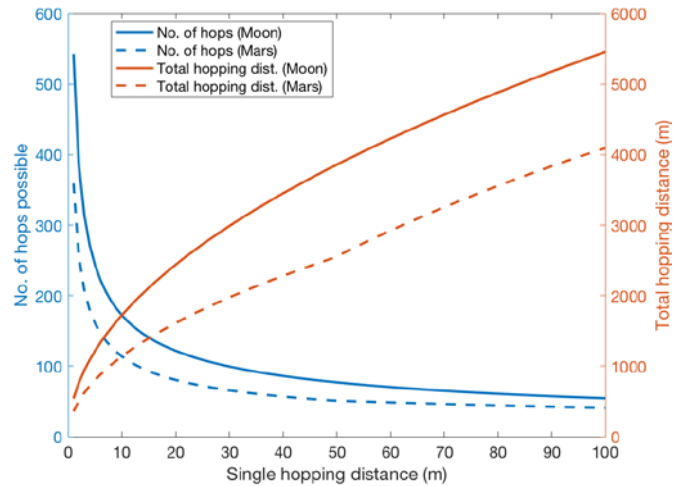

Fig. 3. Number of hops and total hopping distance possible with varrying hopping range in Moon and Mars.

## V. COMMUNICATION SYSTEM

Antenna selection is one of the key factors in the system performance. A trade-off must be made based on antenna radiation pattern and how the antenna affects the structure of the robot. A quad monopole antenna or a dipole antenna could be considered. This has an omnidirectional radiation pattern and will maximize the range of the system in all directions and thus more efficient. However, the system structure is more important to us, hence we have selected a patch antenna. It's radiation pattern is directional and robots in the line of sight of the antenna would receive a stronger signal than other robots and thus, the system would be less efficient. However, this is not a problem as the distances are relatively small..

The communication system in a cave environment is susceptible to multipath and interference, including inter-symbol interference. Hence, an OFDM system with long guard bands is a good implementation to reduce the effect of high multipath [29]. To reduce ISI, an equalizer is implemented [30]. To set up a communication network, individual robots must be identifiable. For this, OFDMA [31] is considered. After this is done, a multi-hop communication network [32] must be set up to create a mesh network of robots [33].

Simulations were performed to estimate antenna range and the transmission time. The antenna link equation is the basis for antenna range calculation. We found the range would be 500 m. The parameters used to calculate the range are shown in Table I.

TABLE I. Parameters for calculation of antenna range

| Parameter | Value |
|---|---|
| Transmitted Power | 25 dBm |
| Antenna Gain | 1 dB |
| Receiver Sensitivity | -80 dBm |
| Frequency | 2.4 GHz |
| Losses | 12 dB |

If a patch antenna is considered, with an 8 dB center frequency gain and beam width around 70 degrees, the least gain in any direction would be at least 1 dB, which is considered for the antenna range calculation. The losses considered include a cable loss of 3 dB and other channel losses due to the cave structure being 9 dB.

Simulations were made to calculate the total time it takes for all the robots to send information to the base robot after multiple hops. The algorithm used to calculate the time taken is as follows:

i. A minimum signal strength required to decode a transmitted signal is calculated using the antenna range calculated in the previous simulation.

ii. An adjacency matrix is then computed based on the current position of all the robots.

iii. A cost function matrix is then computed based on the signal strength.

iv. Dijkstra's algorithm is used to compute the optimum path between all robot nodes, using the adjacency matrix and the cost function matrix.

v. A minimum data rate is calculated for the channel using Shannon's theorem and using the G/T of the receiver, a parameter used to indicate the quality of a receiver.

vi. The total transmission time is calculated based on the number of optimum paths and message transmission time calculated based on the data rate.

Some parameters considered for the simulation are shown in Table II.

TABLE II. Parameters for calculation of transmission time

| Parameter | Value |
|---|---|
| No. of Communication hops | 2-20 |
| Data size | 1 MB |
| System Noise Temperature | 200 K |
| Minimum Eb/No for reception | 10 dB |
| Antenna range | 500 m |
| Channel bandwidth | 20 kHz |
| Pointing loss of antenna | 18 dB |
| Data packet size | 1024 bits |

Fig. 4 shows the time required to transmit 1 MB data based on number of hops. This shows that while we can have a chain of few scores robots to cooperatively pass messages and communicate with a base station, it is not possible extend it further due to the exponential increase in time required. This would enable exploration of a lava tube that is a few hundred to few km in length at a time. For larger lava tubes, the technique would require exploring and fully mapping segments of a lava tube followed by setting up a base station at periodic distances and that would be connected by wire.

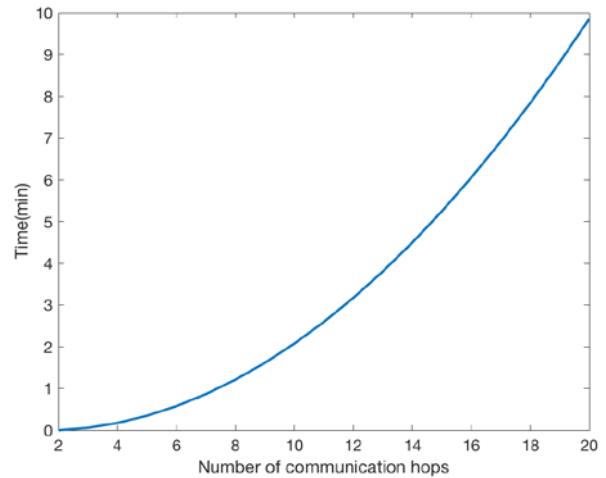

Fig. 4. Time required to transmit 1 MB data through multiple hops.

VI. MULTI-ROBOT PATH PLANNING

In cave environments, there is no line of sight from a starting point to some corridor or cavern. Communication signals are blocked due to rocks in the way. This requires setting up communication relays. Hence, the robots need to cooperate in the form of a bucket brigade to establish a multi-hop communication link as shown in Fig. 5. The communication system has two fixed robots, one at the top of the cave entrance (Base 0) and other at the base of the vertical entrance (Base 1). The Base 0 robot acts as 'base station' that receives data from all the robots inside the cave. The Base 1 robot acts as the intermediary that collects all the information from the other robots inside the cave and transmits it to the Base 0 robot. The remaining robots will perform exploration and at times organize into a bucket brigade establishing a multi-hop communication link from the farthest robot to the Base 0 robot.

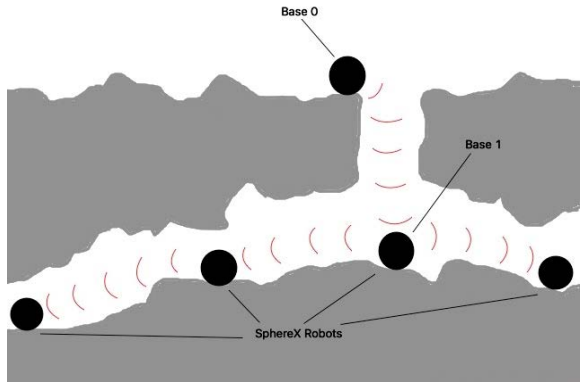

Fig. 5. Multi-hop communication link strategy

The goal of this work is to map a 3D environment of a lava tube using multiple hopping robots, hopping one at a time. The lava tubes of interest are greater than 50 meters in length, and range in width and height between 1 and 8 meters. The tube walls are unpredictable, lacking sharp distinct corners. Using encoder measurements for robot localization is not an option as the robots move by hopping. Low light conditions in the mapping environment cause processing techniques to require structured lighting that may increase payload weight and power consumption. Due to the shielding property of the lava tubes, no radiation communication such as GPS can be established between the robots in the tubes and the outside world [24]. Therefore, a local-based SLAM solution is required.

### A. Robot Position Measurement and Localization

Localization systems such as GPS are not available nor are they easy to install due to signal blockage from the rocks. Each robot is equipped with a 2D laser range finder mounted on a servo to enable 3D range scanning. A global frame ($X, Y$) is constructed w.r.t the fixed robot (Base 1). At any given instant, when one robot hops, its relative position and orientation changes can be measured w.r.t the neighboring stationary robots and then can be converted to the relative position ($x, y$) w.r.t the global frame as shown in Fig. 6.

Robot $i$ hops from its initial position to its final position. Robot $i-1$ measures the range and bearing angle ($R, \alpha$) of Robot $i$ w.r.t its local frame ($X_{i-1}, Y_{i-1}$). The global position ($x_i, y_i$) and orientation $\phi_i$ of Robot $i$ can then be computed w.r.t the global frame ($X, Y$) as shown in (2) and (3). So, if there are $n$ number of robots in between Robot $i$ and Base 1, we must perform $n$ transformation to compute its global position and orientation.

$$\beta = \alpha + \phi_{i-1} \qquad (3)$$

$$\begin{pmatrix} x_i \\ y_i \\ \phi_i \end{pmatrix} = \begin{pmatrix} x_{i-1} \\ y_{i-1} \\ \phi_{i-1} \end{pmatrix} + \begin{pmatrix} R \cos \beta \\ R \sin \beta \\ \theta \end{pmatrix} \qquad (4)$$

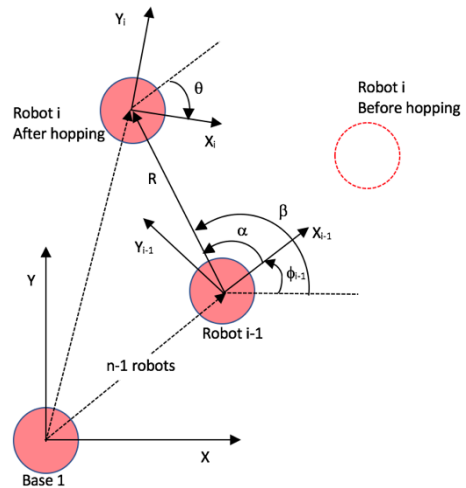

Fig. 6. Robot localization.

### B. Multi-robot algorithm

Exploration of an unknown environment by a multi-robotic system is a well-studied problem but much of the work is confined to wheeled robotic platforms. In this section we present an algorithm for the exploration of an unknown cave environment with the help of multiple hopping robots. In this algorithm, a network of robots uses data from the already explored area for path planning and moves forward one at a time to explore the unknown areas while maintaining communication links with the base station. The multi-robot algorithm presented here is an adaptation of the Sensor-based Random Tree (SRT) navigation tree, modified to explore unknown environments like caves and lava tubes using multiple hopping robots. The cave environment is modeled as a grid with circular obstacles of different size. Each cell in the grid is termed explored once it falls within the sensing radius of any robot. The robot sensors detect its surrounding area and forms a point cloud within the range of its sensors. The exploration is directed through the selection of free boundaries from a given configuration to continue the exploration as shown in Fig. 7(a). Once the free boundary is identified, a random point is selected on the free boundary, and a robot is also selected randomly for hopping. Then a normalized unit vector is computed between the selected point, and the robot identified to hop as shown in Fig. 7(b) and (5).

$$u = \frac{(r_p - r_r)}{\|r_p - r_r\|}$$

Where, $r_p$ is the position vector of the selected point and $r_r$ is the position vector of the robot selected. Once, the normalized vector is computed, the algorithm checks if the route is safe or not which is determined based on the intersection of the unit vector with any obstacles. Fig. 7(b) shows a possible exploration direction since there is a safe path between the two points (solid black line). However, in Fig. 7(d) the exploration direction is not safe due to the presence of an obstacle on its way (dotted red line). Once the exploration direction is termed unsafe, the algorithm selects a different point on the free boundary until a safe path is found. If there is no safe path found, it selects a different robot for hopping. The normalized vector computed determines the hopping direction for the selected robot. Once the hopping direction is determined, the hopping distance must be computed such that the communication graph is connected between the farthest robot and the base robot, the final position must lie within the explored area and it should be within the hopping range of the robot.

With the desired final position computed for the robot selected to hop, the algorithm computes the initial velocity $v_{t0}$, final velocity $v_{tf}$ and the transfer time $\tau$ for its optimal trajectory. The robot hops to its new location and the free boundary is updated as shown in Fig. 7(c). The algorithm then computes the hopping direction and distance for the next robot.

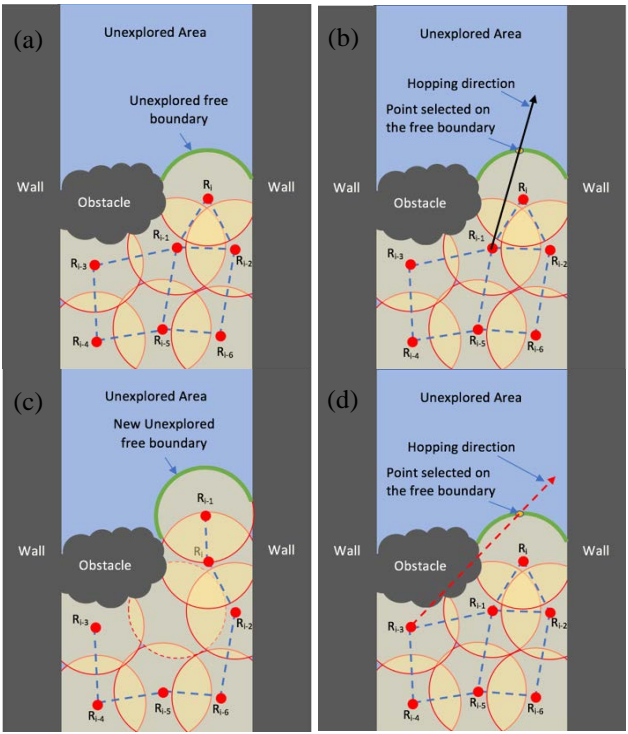

Fig. 7. Path planning and exploration strategy for multiple robots.

In this algorithm, the navigation and exploration are based on the incremental computation of the hopping direction and hopping distance, through the random generation of robot configurations within a local security area detected by the robot sensors where the robots can move freely without the risk of collision with an obstacle.

*C. Algorithm implementation and simulations*

In our method, the environment is built as a grid structure of length 50 units and width 8 units represented by 800x5000 grids consisting of six circular obstacles of different radius. Each robot is considered to have a vision radius of 2 units, communication range of 5 units, and a hopping range of 7 units. Table III. shows the pseudo code for the algorithm. For each iteration k, based on the sensing radius of each robot, the detected grids are declared explored and important features like obstacles are identified. The unexplored free boundary is then computed which determines the hopping direction. After calculation of each hopping direction, verification is done to determine if it's safe or not. Hopping distance is then computed and verified for communication and hopping range constraints. Robot i then hops to its new position and the explored area, obstacles found, and new free boundary are updated. The algorithm then calculates the new positions for each robot based on the updated free boundaries and moves on to the next iteration.

We have considered two cases for the implementation of the algorithm. Case I: The robots should always be in connection/communication with the Base robot. Case II: The robots can explore as a swarm without maintaining connection with the Base robot, collects data and returns back.

TABLE III. Multi-robot path planning pseudo code

| **Algorithm**: Multi-robot path planning for hopping robots |
|---|
| **Require**: Initial position, orientation for each robot; |
| 1.   **for** k = 0 to K **do** |
| 2.     **for** i = 1 to N **do** |
| 3.       Update explored grid cells; |
| 4.       Identify obstacles; |
| 5.       Compute free boundary; |
| 6.       Select random point on free boundary; |
| 7.       Compute hopping direction; |
| 8.       Verify hopping direction; |
| 9.       Compute hopping distance; |
| 10.      Verify hopping distance; |
| 11.      Move robot i to new position; |
| 12.      Update explored grids and obstacles; |
| 13.      Compute new free boundary; |
| 14.      Set i = i+1; |
| 15.    **end for** |
| 16.    Set k = k+1; |
| 17. **end for** |

Case I simulations: Fig. 8 shows multiple SphereX robots explore a lava tube. The red dot denotes the stationary Base robot. The other 15 robots are denoted by black circles and their communication links denoted by solid black line. The unexplored area is denoted by purple, explored area by green and the obstacles by yellow circles.

The robots move forward incrementally within the local known vicinity to identify new features of the cave while leaving robots at strategic locations so that the communication link with the base robot is not broken.

Fig. 9 shows the average percentage of area covered and standard deviation with different number of robots for Case I simulation where a constant link must be maintained with the base robot. The results are obtained based on 10 tests.

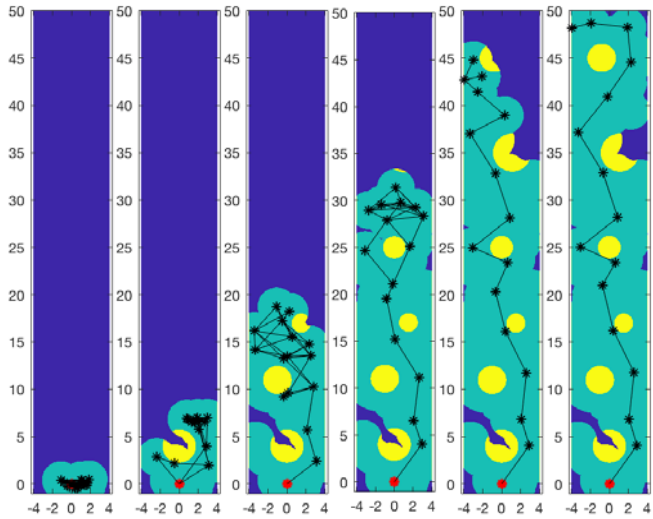

Fig. 8. Case I: Simulation of a system of 15 robots and a base robot at timestep 0, 2, 5, 10, 15 and 20.

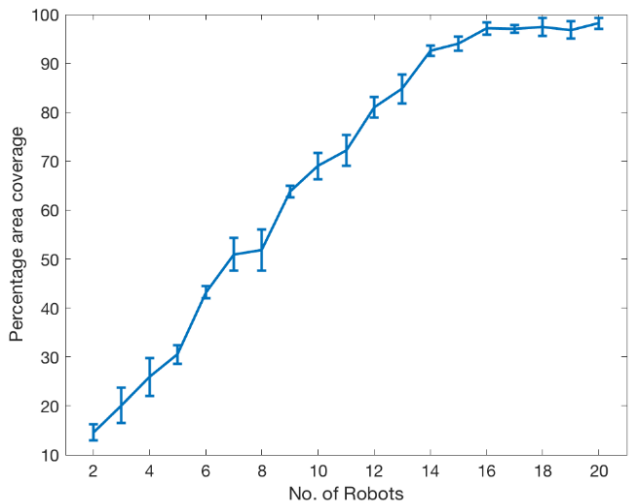

Fig. 9. Average percentage of area covered and standard deviation over 10 tests.

Case II simulations: In case II the robots can break the communication link with the base robot and move forward together maintaining their local communication link. The algorithm works in the same way as for case I except, while computing the hopping distance, it ignores connectivity with the base robot. Fig. 10 shows 6 SphereX robots explore a lava tube.

## VII. DISCUSSION

These simulations show that using a team of SphereX robots it is possible to form a chain-link along the distance of a lava tube. The chain segments the lava tube among the $n$ robots and hence each robot handles mapping, localization and communication in its vicinity. This approach is sufficient to explorer a long stretch of a lava-tube. However, it is unnecessary to have this chain-link stretch all the way from the entrance of a lava-tube to the end. Instead, already mapped segments would need to have base stations to extend access to power, lighting and communications for exploration deeper into the lava-tube.

Credible options include setting up wireless RF base-stations or possibly laser communication links. For power this may include laying wires from a base station outside the lava-tube which is logistically challenging. Alternate, more feasible approaches include setting up a network of mirrors to beam sunlight and laser-light [35] (called the TransFormer strategy [36]) down the lave-tube to recharge robots deep inside the cave. The advantage of beaming light is it solve three problems, namely the lack of light, communication, and power inside the lava-tube.

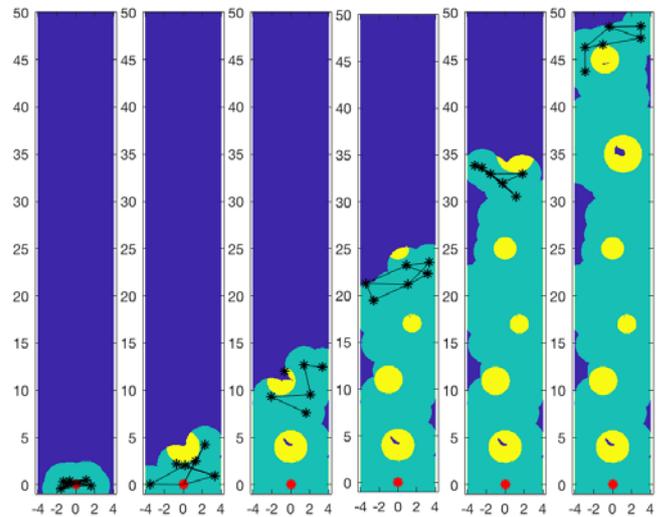

Fig. 10. Case II: Simulation of a system of 6 robots at timestep 0, 2, 6, 11, 15 and 21.

These mirrors equivalent in area to a bathroom tile may well be carried and installed by SphereX robots utilizing its 1 kg payload bay and laid down at candidate relay sites by creating a "trail of breadcrumbs" strategy. This vastly simplifies localization and path planning and can enable the robots exit the lava-tube network quite easily. The robots would simply need to follow along the trail of light to get either deeper into the cave or the entrance.

The use of mirrors to beam light and perform laser communication presents another simplification as there needs to be a straight line of sight from relay to relay. This may work well for lava tubes, where the tube stretches in nearly straight lines several hundred meters at a time. As the distance increase with length $r$, there is indeed a drop off in intensity of $1/r^2$. However, this can be practically solved by diverting more power to the base station at the entrance of the lava tube.

## VIII. CONCLUSION

This paper presented the SphereX robot that uses rocket thrusters to hop, fly and roll in extreme off-word environments such as caves, lava-tubes, and canyons. The proposed concept will allow mapping of these extreme environments compiled into a 3D point cloud using high resolution cameras. They offer the possibility of accessing these sites, never before possible and even performing sample return. Much of the SphereX platform will use COTS hardware. Further development is required in miniaturizing the propulsion system. Moreover, we presented a path planning and navigation algorithm for multiple of these hopping robots in an unknown cave-like environment. The robots hop one at a time and moves deeper into the caves forming a bucket brigade and obtain detailed images. Our high-fidelity computer simulations show the distribution of the robots over time in a grid structured environment. We presented two scenarios. One where the robots need to maintain a constant communication link with the base robot, and the other where the robots can explore leaving the base robot, collect data and return. Efforts are also underway to perform field experiments to validate the simulation results. Our feasibility studies show that with sufficient resources, it is possible to advance the SphereX platform for a technical demonstration in a relevant environment with the future goal of incorporating the robots on a science-led surface mission to the Moon, Mars or asteroids.

ACKNOWLEDGMENT

The authors would like to thank Dr. Mark Robinson from Arizona State University for his invaluable input in the development of the SphereX robot platform.